\documentclass{article}

\usepackage[final]{neurips_2023}
\makeatletter
\renewcommand{\@noticestring}{}
\makeatother

\usepackage[utf8]{inputenc}
\usepackage[T1]{fontenc}
\usepackage{hyperref}
\usepackage{url}
\usepackage{booktabs}
\usepackage{amsfonts}
\usepackage{amsmath}
\usepackage{nicefrac}
\usepackage{microtype}
\usepackage{xcolor}
\usepackage{graphicx}
\usepackage{algorithm}
\usepackage{algorithmic}
\usepackage{enumitem}
\usepackage{tikz}
\usetikzlibrary{arrows.meta, positioning, shapes.geometric, calc, decorations.pathreplacing}
\usepackage{pgfplots}
\pgfplotsset{compat=1.18}
\usepgfplotslibrary{groupplots}
\usepackage{listings}
\usepackage{multirow}
\usepackage{pifont}
\newcommand{\cmark}{\ding{51}}

\title{StreamIndex: Memory-Bounded Compressed Sparse Attention via Streaming Top-k}

\author{
  Jaber Jaber\thanks{Correspondence: \texttt{jaber@rightnowai.co}} \\
  RightNow AI\\
  \texttt{jaber@rightnowai.co} \\
  \And
  Osama Jaber \\
  RightNow AI\\
  \texttt{osama@rightnowai.co} \\
}

\begin{document}
\maketitle

\begin{center}
\includegraphics[height=1.1cm]{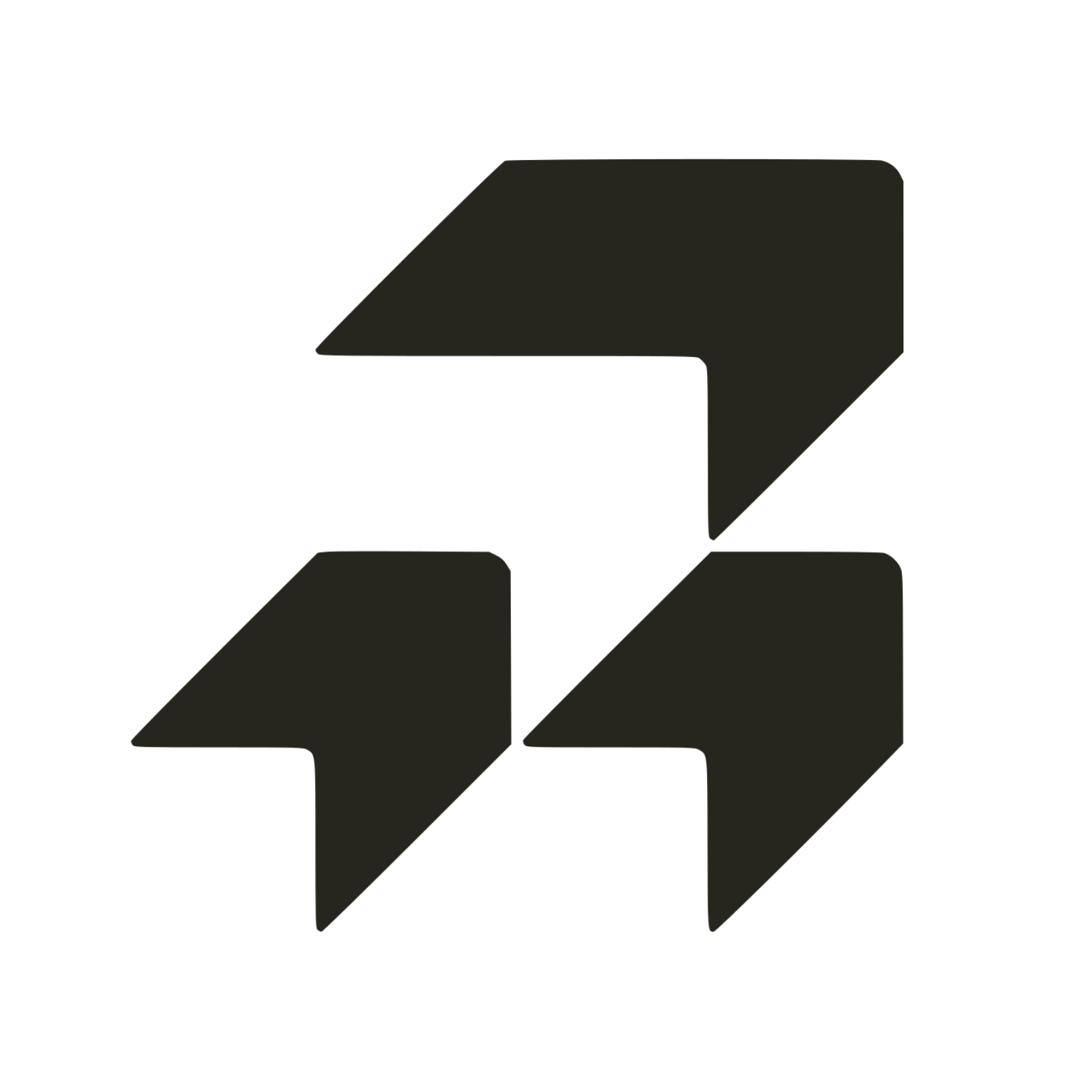}
\end{center}
\vspace{-0.3em}

\begin{abstract}
DeepSeek-V3.2 and V4 introduce Compressed Sparse Attention (CSA): a lightning indexer (a learned scoring projection over compressed keys) scores them, the top-$k$ are selected per query, and a sparse attention kernel reads only those. Public CSA implementations materialize a $[B, S, H_I, T]$ FP32 score tensor before the top-$k$ reduction. With $H_I{=}64$ indexer heads and the V4-Flash compression ratio $m{=}4$, that intermediate is 256~GB at sequence length $S{=}65{,}536$, exceeding any single-GPU high-bandwidth-memory (HBM) budget. We present \textsc{StreamIndex}, a Triton implementation of the CSA pipeline whose central component is a chunked partition-merge top-$k$ driver that never materializes the full intermediate. On synthetic-but-realistic V4-shaped inputs at the indexer-step (layer) level on a single NVIDIA H200, the materialize path runs out of memory (OOMs) at $S{=}65{,}536$ with V4-Flash dimensions; \textsc{StreamIndex} runs the same indexer to $S{=}1{,}048{,}576$ with 6.21~GB peak HBM, a 32$\times$ regime extension. Set-overlap recall against the materialize ground truth is bit-exact at small $S$ where both fit; across three 5-point design-space sweeps (chunk size, key-tile size, top-$k$), mean recall rounds to 1.0000 with min recall at least 0.9980 in every cell. The chunked driver composes with TileLang's pipelined attention kernel: at $S{=}262{,}144$ with V4-Flash dimensions, the materialize indexer paired with TileLang attention OOMs while the chunked indexer paired with the same attention runs in 1.97~s at 18.56~GB peak. Our contribution targets the indexer step; we make no claim of a faster attention kernel or of real-checkpoint end-to-end behavior. Code: \url{https://github.com/RightNow-AI/StreamIndex}.
\end{abstract}

\section{Introduction}

DeepSeek-V3.2 and V4 attach a learned indexer to the attention path~\cite{deepseekv32}. The indexer scores compressed keys, the top-$k$ are selected, and a sparse attention kernel reads only those keys for each query. The pattern preserves quality at long context while reducing the attention step from $O(S^2)$ to $O(S \cdot k)$. V4-Flash and V4-Pro extend this to a 1M-token context with $H_I{=}64$ indexer heads, $m{=}4$ key compression, and per-query top-$k \in \{512, 1024\}$~\cite{deepseekv32}.

The claim that CSA scales to 1M context relies on the indexer step itself fitting. The lightning-indexer score
$I(t, s) = \sum_{h=1}^{H_I} w_{t,h} \cdot \mathrm{ReLU}(q_{t,h} \cdot K_s)$
is computed in the V4 reference (\texttt{model.py}) and in TileLang's reference \cite{tilelang} as a fused einsum producing the intermediate $[B, S, H_I, T]$ FP32 tensor before the head sum. With V4-Flash dimensions ($H_I{=}64$, $T{=}S/m$), this intermediate is 256~GB at $S{=}65{,}536$ and 1~TB at $S{=}131{,}072$ on a single batch. Every public CSA pipeline we are aware of OOMs on a single H200 (140~GB HBM) for $S \geq 64$K with V4-Flash dimensions, before the top-$k$ has been selected, before any attention has run.

Existing long-context attention work attacks the attention step directly: FlashAttention~\cite{flashattn1,flashattn2,flashattn3} fuses softmax with the matmul, paged attention~\cite{vllm} chunks the KV cache, ring attention~\cite{ringattn} sequence-parallelizes. None of these address the indexer-step intermediate that gates V4-class CSA at long context. Native Sparse Attention (NSA)~\cite{nsa} replaces the indexer entirely with a learned top-$k$ structure, but inherits a different memory profile and is not drop-in compatible with V4 checkpoints.

We attack the indexer step directly. The lightning-indexer score is per-key separable: once the query is fixed, the score is a pure function of the key index, with no inter-key coupling (no softmax, no normalization that mixes scores). This separability makes the top-$k$ commute with iteration order over the key axis, which is folklore in streaming algorithms~\cite{charikar}. We apply that fact to the V4 indexer in a chunked partition-merge implementation: process queries and keys in $(c_S, c_T)$ tiles, take per-tile top-$k$, merge across tiles. Peak HBM for the score buffer drops from $O(B \cdot S \cdot H_I \cdot T)$ to $O(B \cdot c_S \cdot c_T)$.

\paragraph{Key insight.} The indexer step is what gates CSA at long context, not the attention kernel. The attention step has streaming patterns (FlashAttention) and is well-served by existing CUDA kernels; the indexer-step intermediate is the unaddressed bottleneck. A Python-level chunked driver over a fused score kernel removes the peak-memory bottleneck while empirically matching the materialize top-$k$ as a set on tested parity cases.

\paragraph{Contributions.}
\begin{enumerate}[leftmargin=*,topsep=2pt,itemsep=2pt]
\item \textsc{StreamIndex}, an open-source Triton implementation of the V4 CSA pipeline whose chunked partition-merge top-$k$ driver realizes the streaming top-$k$ invariance for the lightning indexer (151 lines, \texttt{chunked\_indexer.py}).
\item A per-chunk causal-mask formulation that keeps mask memory at $O(c_S \cdot c_T)$ instead of $O(S \cdot T)$, which is what allows the chunked path itself to scale past $S{=}131{,}072$.
\item A measured 32$\times$ regime extension at V4-Flash dimensions on a single H200: the materialize indexer OOMs at $S{=}65{,}536$, our chunked indexer runs at $S{=}1{,}048{,}576$ with 6.21~GB peak HBM.
\item A pipeline-composition result: at $S{=}262{,}144$ with V4-Flash dimensions and TileLang's pipelined attention kernel as the common backend, the materialize-indexer path OOMs while our chunked-indexer path runs at 1.97~s, 18.56~GB peak.
\item A design-space and ablation study (3 sweeps, 3 ablations) showing that chunk size has a clear knee around $c_S{=}2048$, single-chunk over $T$ is optimal when memory allows, and FP32 score accumulation is necessary (FP16 drops perfect-recall rows from 99.99\% to 91.82\%).
\item Bit-exact set-overlap recall versus the V4-Flash materialize-then-postmask reference at $S \in \{2048, 4096, 8192\}$, validated under realistic post-projection input distributions.
\end{enumerate}

\section{Related Work}

\begin{table}[t]
\centering
\caption{Long-context attention systems and the bottleneck each addresses. The CSA indexer-step intermediate ($O(B \cdot S \cdot H_I \cdot T)$) is unaddressed by every prior system; \textsc{StreamIndex} is the only entry whose primary target is the indexer.}
\label{tab:relwork}
\scriptsize
\setlength{\tabcolsep}{3pt}
\renewcommand{\arraystretch}{1.05}
\begin{tabular}{@{}p{2.7cm}p{2.0cm}p{3.0cm}p{2.7cm}c@{}}
\toprule
\textbf{System} & \textbf{Target step} & \textbf{Approach} & \textbf{Sparsity} & \textbf{OSS} \\
\midrule
FlashAttention 2/3~\cite{flashattn2,flashattn3} & attention & fused softmax+matmul & dense & \cmark \\
PagedAttention (vLLM)~\cite{vllm} & KV cache & block paging & dense & \cmark \\
Ring Attention~\cite{ringattn} & attention & sequence parallel & dense & \cmark \\
StreamingLLM~\cite{streamingllm} & KV cache & sliding window + sinks & fixed pattern & \cmark \\
H2O~\cite{h2o} & KV cache & heavy-hitter eviction & learned pattern & \cmark \\
Quest~\cite{quest} & attention & query-aware page select & query-dependent & \cmark \\
MInference~\cite{minference} & attention & dynamic head patterns & dynamic & \cmark \\
NSA~\cite{nsa} & attention + indexer & trained sparsity & learned top-$k$ & \cmark \\
V4 reference~\cite{deepseekv32} & indexer + attention & materialize + topk & top-$k$ over compressed & \cmark$^{\dagger}$ \\
\textbf{\textsc{StreamIndex}} & \textbf{indexer} & \textbf{chunked partition-merge} & \textbf{top-$k$ over compressed} & \cmark \\
\bottomrule
\end{tabular}
\par\vspace{2pt}
{\scriptsize $^{\dagger}$\,V4 / V3.2 reference materializes the $[B, S, H_I, T]$ FP32 intermediate.}
\end{table}

Table~\ref{tab:relwork} positions \textsc{StreamIndex} against prior long-context attention work. Every entry except the V4 reference and \textsc{StreamIndex} optimizes the attention or KV-cache step. \textsc{StreamIndex} is the first open-source CSA implementation that does not materialize the indexer-step intermediate.

\paragraph{Sparse and compressed attention.}
The structured-sparsity line started with Sparse Transformers~\cite{sparsetransformers}, Longformer~\cite{longformer}, and BigBird~\cite{bigbird}: predetermined block, dilated, or random sparsity patterns. DeepSeek-V2~\cite{deepseekv2} and V3~\cite{deepseekv3} introduced Multi-Latent Attention (MLA) with a low-rank KV factorization. V3.2~\cite{deepseekv32} adds the Compressed Sparse Attention pattern with the lightning indexer; V4-Flash and V4-Pro generalize to 1M context. Native Sparse Attention~\cite{nsa} learns a sparsity pattern jointly with the model. None of the public reference kernels for these methods address the indexer-step memory peak we attack.

\paragraph{Query-aware sparse inference.}
Quest~\cite{quest} and MInference~\cite{minference} both compute query-dependent sparsity at inference time. Quest selects KV pages by a small score per page; MInference picks dynamic patterns per attention head. Both target the attention-step compute, not the indexer-step intermediate. Selective Attention~\cite{selectiveattn} provides a related principled mechanism. Mixture-of-experts work~\cite{mixtral} influences the broader V4 architecture but is orthogonal to the attention path.

\paragraph{IO-aware attention.}
FlashAttention~\cite{flashattn1,flashattn2,flashattn3} fuses softmax with QK and PV matmuls so attention is computed without materializing the $S \times S$ score matrix. The technique applies to dense attention; in CSA it bypasses the attention-step bottleneck but leaves the indexer-step intermediate intact.

\paragraph{Long-context infrastructure.}
PagedAttention in vLLM~\cite{vllm} chunks the KV cache. Ring attention~\cite{ringattn} sequence-parallelizes across devices. StreamingLLM~\cite{streamingllm} keeps a fixed window plus attention sinks. H2O~\cite{h2o} retains heavy hitters. Each addresses the KV-cache or attention pass; the V4 indexer's $[B, S, H_I, T]$ intermediate is orthogonal.

\paragraph{Approximate and structured attention.}
Linformer~\cite{linformer}, Performer~\cite{performer}, Reformer~\cite{reformer}, and linear attention~\cite{linearattn} approximate softmax attention with low-rank, locality-sensitive-hashing (LSH), or kernel structure. State-space models such as Mamba~\cite{mamba} and Hyena~\cite{hyena} avoid attention entirely. CSA, which V4-Flash uses, is exact (top-$k$ keys are read fully); we preserve that property. Multi-query attention~\cite{mqa} and its variants reduce KV-cache size by sharing keys/values across heads; V4-Flash uses an MLA layout that is a generalization.

\paragraph{Streaming top-k.}
Partition-merge invariance for top-$k$ over a separable scoring function is folklore; Charikar et al.~\cite{charikar} state and use it for streaming heavy-hitter problems, and the textbook treatment is in Mitzenmacher and Upfal~\cite{mitzenmacher}. Our contribution is engineering: applying it to the V4 lightning indexer in the first publicly-available CSA implementation we are aware of that does not materialize the score intermediate.

\paragraph{Triton and tiled kernels.}
Triton~\cite{triton} is the kernel language we use. TileLang~\cite{tilelang} provides reference CSA kernels that we benchmark against (their attention kernel; their reference indexer materializes).

\section{Background: V4 Compressed Sparse Attention}

\paragraph{The CSA pipeline.}
A V4 CSA layer takes a hidden state $x \in \mathbb{R}^{B \times S \times d}$ and produces an attention output of the same shape. Internally:
\begin{enumerate}[leftmargin=*,topsep=1pt,itemsep=1pt]
\item A token compressor produces a compressed key/value cache $K_C \in \mathbb{R}^{B \times T \times d_h}$ where $T = S/m$ ($m{=}4$ for V4-Flash).
\item A lightning indexer scores each compressed key for each query: $I(t, s) = \sum_h w_{t,h} \mathrm{ReLU}(q_{t,h} \cdot K_C^s)$ for $h \in [H_I]$, $s \in [T]$, $t \in [S]$.
\item For each query $t$, the indexer selects $\mathrm{TopK}(t)$, the indices of the top-$k$ legal compressed keys.
\item A sparse attention kernel reads only $\mathrm{TopK}(t) \cup \mathrm{Window}(t)$ for each query and returns the attention output.
\end{enumerate}

\paragraph{The indexer-step intermediate.}
The reference V4-Flash forward (\texttt{references/DeepSeek-V4-Flash/inference/model.py}, lines 415--423) computes the indexer score as
\begin{lstlisting}
index_score = torch.einsum("bshd,btd->bsht", q, kv_cache_slice)
index_score = (index_score.relu_() * weights.unsqueeze(-1)).sum(dim=2)
\end{lstlisting}
The first line produces a $[B, S, H_I, T]$ FP32 tensor; the second reduces over $H_I$. With V4-Flash dimensions ($B{=}1$, $H_I{=}64$, $T{=}S/4$), this intermediate is $256$~GB at $S{=}65{,}536$ and 4~TB at $S{=}262{,}144$.

This is the bottleneck. The downstream attention kernel does not see it. The Triton-based or TileLang-based attention kernel that comes after has access only to the compact top-$k$ index list. The score intermediate is allocated and freed inside the indexer step alone, which means {\it the indexer step gates the maximum sequence length the pipeline can run, regardless of how efficient the attention kernel is}. Assuming attention is the bottleneck at long context is incorrect for CSA: the pipeline runs out of memory in the indexer step before attention runs.

\paragraph{Per-element variance.}
The lightning-indexer score is well-conditioned at training scale: $q$ is rotary-position-embedded~\cite{rope}, Hadamard-rotated~\cite{quarot}, and quantized to FP4~\cite{microscaling}, so $q_{t,h} \cdot K_C^s$ has approximately unit variance per element. Weights $w_{t,h}$ are unconstrained-sign output of a trained linear projection. We use these distributional facts to construct synthetic but spec-matched inputs for the parity tests in \S\ref{sec:eval}.

\section{Streaming Top-k for the Lightning Indexer}
\label{sec:math}

\subsection{Total order on (score, index) pairs}
Top-$k$ requires a tie-breaking rule. We define $\succ$ on $\mathbb{R} \times \mathbb{Z}_{\geq 0}$ by
\[
(a, i) \succ (b, j) \quad \Longleftrightarrow \quad a > b \;\lor\; (a = b \;\land\; i < j).
\]
This is a strict total order. Ties on score are broken in favor of the smaller index, matching the inductive bias of causal language modeling. PyTorch's \texttt{torch.topk} does not guarantee any tie-break, so set comparison (not index-position comparison) is the right notion of equivalence.

\subsection{Effective top-k}
Causal masking restricts the legal index range. Let $T_{\mathrm{legal}}(t) = \lfloor (t+1)/m \rfloor$ be the number of compressed blocks fully inside the past of query $t$. The output size is $k_{\mathrm{eff}}(t) = \min(k, T_{\mathrm{legal}}(t))$, padded to $k$ with the sentinel value $-1$ when $T_{\mathrm{legal}}(t) < k$.

\subsection{The partition-merge invariance}
\begin{algorithm}[t]
\caption{Streaming top-$k$ for one query}
\label{alg:streaming}
\begin{algorithmic}[1]
\REQUIRE Score function $g : [T_{\mathrm{legal}}] \to \mathbb{R}$, target $k$
\STATE $H \gets \emptyset$ \quad ({\tt$\succ$-min heap of size $\le k$})
\FOR{each $s$ in any permutation of $[T_{\mathrm{legal}}]$}
\STATE $\sigma \gets g(s)$
\IF{$|H| < k$}
\STATE insert $(\sigma, s)$ into $H$
\ELSIF{$(\sigma, s) \succ H.\mathrm{peek}()$}
\STATE pop the $\succ$-min from $H$; insert $(\sigma, s)$
\ENDIF
\ENDFOR
\RETURN $H$
\end{algorithmic}
\end{algorithm}

\paragraph{Theorem (partition-merge invariance, idealized deterministic top-$k$).}
\textit{Algorithm~\ref{alg:streaming} returns the unordered set of the top-$\min(k, T_{\mathrm{legal}})$ pairs under $\succ$, regardless of the permutation chosen in the loop. We write this set as $\mathrm{argtop}_{\succ}^k g$. Sentinel padding (used when $T_{\mathrm{legal}} < k$) is post-processing and is excluded from set comparisons.}

\noindent\textit{Proof sketch.} Strong induction on prefix length $j$. Let $M_j$ be the top-$k$ of the first $j$ processed pairs under $\succ$, and $H_j$ the heap state after step $j$. The base case $j{=}0$ gives $M_0 = H_0 = \emptyset$. The inductive step splits on whether $j{+}1 \le k$ (every element survives) or $j{+}1 > k$. In the latter case, the new element either is $\succ$-greater than $H_j.\mathrm{peek}()$ (it displaces the $\succ$-min, preserving top-$k$) or not (it does not enter $M_{j+1}$, again preserving top-$k$). In all cases $H_{j+1} = M_{j+1}$. The full proof, with the tied-score case, is in the source repository at \texttt{docs/streaming\_topk.md}. $\square$

\paragraph{Implementation note (set parity, not order parity).} The strict total order $\succ$ is defined for the formal theorem statement, but our implementation uses \texttt{torch.topk} on raw FP32 scores, whose tie-breaking is unspecified. We therefore do not derive set-parity from the $\succ$-theorem; we \emph{empirically} verify (\S\ref{sec:eval}) bit-exact set match between the materialize and chunked top-$k$ outputs. On all parity tests at V4-Flash dimensions the two paths agree on the set, including under FP32 tied-score conditions. This empirical guarantee matches the practical use of the indexer, which feeds the index set to the attention kernel where order is irrelevant; promoting it to a derived corollary would require inlining a deterministic comparator in both paths, which neither the materialize reference nor our implementation does.

\paragraph{Corollary (chunked partition-merge).}
For any partition $[T_{\mathrm{legal}}] = P_1 \sqcup P_2 \sqcup \dots \sqcup P_n$,
\[
\mathrm{argtop}^k_{\succ} g = \mathrm{topk}_{\succ}\Bigl(\bigcup_{i=1}^{n} \mathrm{argtop}^{\min(k, |P_i|)}_{\succ} g|_{P_i}\Bigr).
\]
This is what enables chunked execution: process partitions independently, take per-partition top-$k$, merge.

\section{The Chunked Indexer}
\label{sec:design}

\begin{figure}[t]
\centering
\begin{tikzpicture}[
  box/.style={draw, rounded corners=3pt, thick, minimum height=0.6cm,
              align=center, font=\footnotesize\sffamily, text width=6.0cm},
  arr/.style={-{Stealth[length=5pt]}, thick},
  loopback/.style={-{Stealth[length=5pt]}, thick, dashed, gray!70!black},
  node distance=0.32cm,
]
\node[box, fill=blue!10] (q) {hidden state $x \in \mathbb{R}^{B \times S \times d}$};
\node[box, fill=blue!10, below=of q] (proj) {linear projections\\$q{:}\,[B,S,H_I,d_h]\quad K_C{:}\,[B,T,d_h]\quad w{:}\,[B,S,H_I]$};
\node[box, fill=orange!15, below=of proj] (driver) {\textbf{chunked driver} (\texttt{chunked\_indexer.py}, 151 LOC)\\[1pt]
  \scriptsize for each $(c_S, c_T)$ tile: dispatch $\to$ score $\to$ merge};
\node[box, fill=red!10, below=of driver] (score) {fused score kernel (\texttt{indexer\_score.py}, 198 LOC)\\[1pt]
  \scriptsize $\mathrm{score} = \sum_h w_h \cdot \mathrm{ReLU}(q_h \cdot K_C^{\,t})$ on $[c_S, c_T]$ tile};
\node[box, fill=red!10, below=of score] (merge) {top-$k$ + merge into running buffer of size $k$};
\node[box, fill=blue!10, below=of merge] (out) {top-$k$ indices $\in \mathbb{Z}^{B \times S \times k}$};

\draw[arr] (q) -- (proj);
\draw[arr] (proj) -- (driver);
\draw[arr] (driver) -- node[right, font=\scriptsize, midway] {tile} (score);
\draw[arr] (score) -- (merge);
\draw[arr] (merge) -- (out);
\draw[loopback]
  (merge.west) -- ++(-1.0,0)
  |- node[left, font=\scriptsize, pos=0.25, align=left] {next\\$(c_S, c_T)$} (driver.west);
\end{tikzpicture}
\caption{The chunked indexer pipeline. The driver iterates over $(c_S, c_T)$ tiles in nested loops over $S$ and $T$; each tile dispatch produces a small $[c_S, c_T]$ FP32 score buffer that is consumed and merged before the next tile. Peak HBM for the score buffer is $O(B \cdot c_S \cdot c_T)$, independent of total $S$ and $T$. The dashed gray edge shows the per-tile loop.}
\label{fig:arch}
\end{figure}
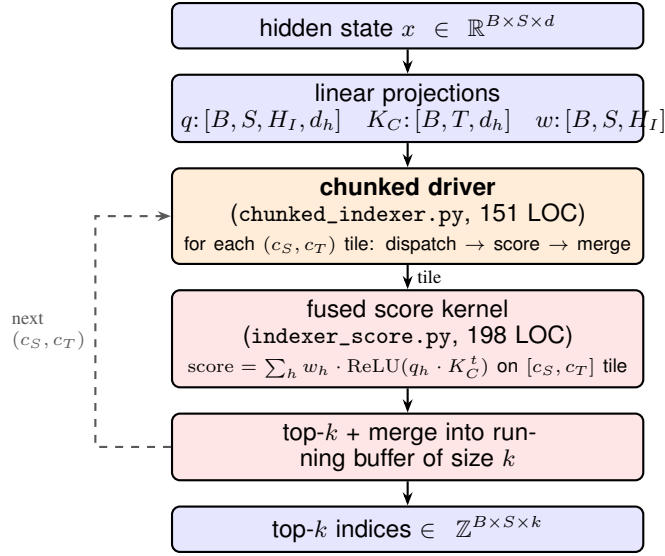

\textsc{StreamIndex} instantiates the corollary of \S\ref{sec:math} as a Python driver over Triton primitives. Figure~\ref{fig:arch} shows the data flow.

\paragraph{Fused score kernel (\texttt{indexer\_score.py}, 198 lines).}
A single autotuned Triton kernel computes
$\mathrm{score}_{b,t,s} = \sum_h w_{b,t,h} \cdot \mathrm{ReLU}(q_{b,t,h,d} \cdot K_{C\,b,s,d})$
on a $[c_S, c_T]$ tile. The tensor of head-wise scores is reduced to $[c_S, c_T]$ FP32 inside the kernel before any global memory write; we never materialize the $H_I$-axis intermediate. This is the unit of work the chunked driver issues per tile.

\paragraph{Chunked driver (\texttt{chunked\_indexer.py}, 151 lines).}
The driver is the body of Algorithm~\ref{alg:driver}. It maintains a running top-$k$ buffer per query, indexed by the outer $S$-tile. For each $T$-tile, it calls the fused score kernel, applies the per-tile causal mask (\S\ref{para:causal}), takes per-tile top-$k$, and merges into the running buffer.

\begin{algorithm}[t]
\caption{Chunked indexer top-$k$}
\label{alg:driver}
\begin{algorithmic}[1]
\REQUIRE $q \in [B, S, H_I, d_h]$, $K_C \in [B, T, d_h]$, $w \in [B, S, H_I]$, target $k$, ratio $m$, tile sizes $c_S, c_T$
\STATE $\mathrm{out} \gets \mathtt{full}([B, S, k], -1)$
\FOR{$s_0 \in [0, S, c_S)$}
\STATE $q_S, w_S \gets q[\,:,s_0:s_0+c_S], w[\,:,s_0:s_0+c_S]$
\STATE $\mathrm{run\_v} \gets \mathtt{full}([B, c_S, k], -\infty)$;
       $\mathrm{run\_i} \gets \mathtt{full}([B, c_S, k], -1)$
\FOR{$t_0 \in [0, T, c_T)$}
\STATE $K_t \gets K_C[\,:,t_0:t_0+c_T]$
\STATE $\mathrm{scores} \gets \mathtt{indexer\_score}(q_S, K_t, w_S)$ \quad ({\tt$[B, c_S, c_T]$ FP32})
\STATE apply per-tile causal mask using $\mathrm{ratio}{=}m$ and $(s_0, t_0)$ offsets
\STATE $(v, i) \gets \mathrm{scores.topk}(\min(k, c_T))$
\STATE $\mathrm{run\_v}, \mathrm{run\_i} \gets \mathrm{merge\_topk}(\mathrm{run\_v}, \mathrm{run\_i}, v, i+t_0)$
\ENDFOR
\STATE $\mathrm{out}[:, s_0:s_0+c_S] \gets \mathrm{run\_i}$
\ENDFOR
\RETURN $\mathrm{out}$
\end{algorithmic}
\end{algorithm}

\paragraph{Per-chunk causal mask.}
\label{para:causal}
A naive implementation passes a $[B, S, T]$ bool causal mask. At $S{=}131{,}072$, $T{=}32{,}768$ that mask is 4~GB; at $S{=}1$M, $T{=}256$K it is 256~GB. We add an integer parameter \texttt{causal\_ratio} to the chunked driver: the mask
$t < \lfloor (s+1)/m \rfloor$ is constructed per $(c_S, c_T)$ tile from \texttt{torch.arange} on the appropriate offsets. Per-tile mask memory is $O(c_S \cdot c_T)$, independent of $S$ and $T$. This single change is what allows the chunked path itself to scale past $S{=}131{,}072$.

\paragraph{Auto-detection.}
The end-to-end CSA forward (\texttt{flash\_sparse/csa.py}, 199 lines) selects the materialize path when the would-be score matrix fits in 1~GB and the chunked path otherwise. Small-$S$ workloads stay on the materialize path, which is faster when it fits (no merge overhead).

\section{Experimental Evaluation}
\label{sec:eval}

We measure \textsc{StreamIndex}'s chunked indexer (\S\ref{sec:design}) along five axes: bit-exact parity at small $S$, layer-level scaling and the OOM threshold (\S\ref{ssec:scaling}), V4-Pro dimensions, full-pipeline composition with TileLang attention, and design-space sweeps and ablations.

\paragraph{Hardware and methodology.}
All runs are on a single NVIDIA H200 SXM with 140~GB HBM3e, BF16 precision, CUDA 13, Triton 3.7, PyTorch 2.13 nightly. Timings use \texttt{torch.cuda.Event} with at least 1 warmup and 3 measured iterations after autotune. Peak HBM is measured with \texttt{torch.cuda.max\_memory\_allocated}, with baseline subtracted after a separate warmup pass so JIT compilation is excluded.

\paragraph{Synthetic-but-realistic inputs.}
Where the V4-Flash modelling-code projections cannot be loaded (no checkpoint on a single H200 fits the 270~GB FP8~\cite{fp8} V4-Flash weights), we draw $q, K_C \sim \mathcal{N}(0, 1/d_h)$ to match per-element variance after the wq\_b $\to$ unflatten $\to$ RoPE~\cite{rope} $\to$ Hadamard~\cite{quarot} $\to$ FP4~\cite{microscaling} pipeline, and use a freshly-initialized \texttt{nn.Linear($d, H_I$)} for the weights projection scaled by $\frac{1}{\sqrt{d_h \cdot H_I}}$ to match the V4 reference. We label all numerical results in this section as ``synthetic-but-realistic.''

\subsection{Bit-exact parity at small \texorpdfstring{$S$}{S}}

We replicate the V4-Flash \texttt{Indexer.forward} materialize+topk+postmask code path verbatim and compare its output to our chunked-indexer output via per-query set-overlap recall over valid (non-pad) entries, with a strict gate (mean $=$ min $=$ 1.0).

\begin{table}[t]
\centering
\caption{V4-Flash indexer parity test, $H_I{=}64$, $d_h{=}128$, $k{=}512$, ratio$=4$. Pass criterion: bit-exact set match per row.}
\label{tab:parity}
\footnotesize
\setlength{\tabcolsep}{4pt}
\begin{tabular}{@{}rrrrrrr@{}}
\toprule
$S$ & $T$ & mean rec. & min rec. & \% rows = 1 & \% rows $<$ .99 & verdict \\
\midrule
2{,}048 & 512 & 1.0000 & 1.0000 & 100.00 & 0.000 & \textbf{PASS} \\
4{,}096 & 1{,}024 & 1.0000 & 1.0000 & 100.00 & 0.000 & \textbf{PASS} \\
8{,}192 & 2{,}048 & 1.0000 & 1.0000 & 100.00 & 0.000 & \textbf{PASS} \\
\bottomrule
\end{tabular}
\end{table}

Table~\ref{tab:parity} reports 100\% mean and min recall on every tested row. The chunked indexer is bit-exact on this workload.

\subsection{Layer-level scaling: 32\texorpdfstring{$\times$}{x} regime extension}
\label{ssec:scaling}

\begin{table}[t]
\centering
\caption{V4-Flash indexer S-scaling on a single H200. Materialize path is the V4-Flash reference; chunked path uses $c_S{=}2048, c_T{=}8192$. ``HBM'' is peak working set excluding inputs. Speedup is materialize-time / chunked-time when both succeed.}
\label{tab:scaling}
\footnotesize
\setlength{\tabcolsep}{3pt}
\begin{tabular}{@{}rrrrrrr@{}}
\toprule
$S$ & $T$ & mat (ms) & mat HBM & chunk (ms) & chunk HBM & speedup \\
\midrule
32{,}768 & 8{,}192 & 317.0 & 129.00 GB & \textbf{31} & \textbf{0.40 GB} & 10.3$\times$ \\
65{,}536 & 16{,}384 & \textbf{OOM} & --- & 122 & 0.59 GB & $\infty$ \\
131{,}072 & 32{,}768 & OOM & --- & 485 & 0.96 GB & $\infty$ \\
262{,}144 & 65{,}536 & OOM & --- & 1{,}935 & 1.71 GB & $\infty$ \\
524{,}288 & 131{,}072 & OOM & --- & 7{,}730 & 3.21 GB & $\infty$ \\
\textbf{1{,}048{,}576} & \textbf{262{,}144} & \textbf{OOM} & --- & \textbf{30{,}900} & \textbf{6.21 GB} & $\boldsymbol{\infty}$ \\
\bottomrule
\end{tabular}
\end{table}

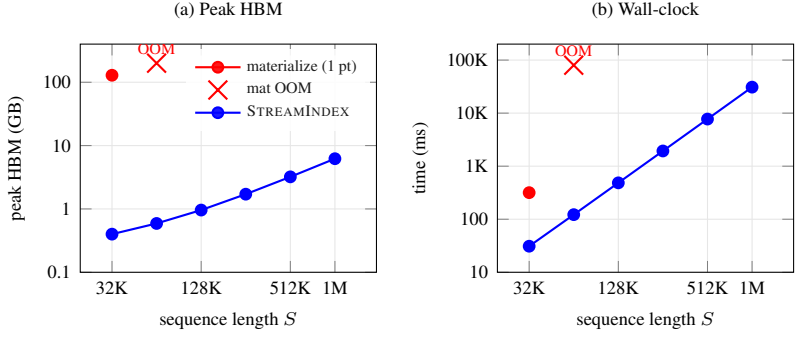
\begin{figure}[t]
\centering
\begin{tikzpicture}
\begin{groupplot}[
  group style={group size=2 by 1, horizontal sep=1.6cm},
  width=5.5cm, height=4.6cm,
  xmode=log, log basis x=2, xlabel={sequence length $S$},
  xmin=20000, xmax=2000000,
  xtick={32768,131072,524288,1048576},
  xticklabels={32K,128K,512K,1M},
  xticklabel style={font=\scriptsize}, yticklabel style={font=\scriptsize},
  label style={font=\scriptsize}, title style={font=\scriptsize},
  legend style={font=\tiny, draw=none, fill=white, fill opacity=0.9,
                at={(0.98,0.98)}, anchor=north east, cells={anchor=west},
                row sep=-1pt},
  grid=both, grid style={gray!20},
]
%--- Panel (a) ---
\nextgroupplot[ymode=log, log basis y=10, ylabel={peak HBM (GB)},
               title={(a) Peak HBM},
               ymin=0.1, ymax=400,
               ytick={0.1,1,10,100}, yticklabels={0.1,1,10,100}]
\addplot[mark=*, thick, red, mark size=2pt] coordinates {(32768,129.0)};
\addplot[mark=x, thick, red, mark size=5pt, only marks] coordinates {(65536,200)};
\addplot[mark=*, thick, blue, mark size=2pt] coordinates {(32768,0.40) (65536,0.59) (131072,0.96) (262144,1.71) (524288,3.21) (1048576,6.21)};
\node[font=\tiny, red, anchor=south] at (axis cs:65536,200) {OOM};
\legend{materialize (1 pt), mat OOM, \textsc{StreamIndex}}
%--- Panel (b) ---
\nextgroupplot[ymode=log, log basis y=10, ylabel={time (ms)},
               title={(b) Wall-clock},
               ymin=10, ymax=200000,
               ytick={10,100,1000,10000,100000}, yticklabels={10,100,1K,10K,100K}]
\addplot[mark=*, thick, red, mark size=2pt] coordinates {(32768,317.0)};
\addplot[mark=x, thick, red, mark size=5pt, only marks] coordinates {(65536,80000)};
\addplot[mark=*, thick, blue, mark size=2pt] coordinates {(32768,31) (65536,122) (131072,485) (262144,1935) (524288,7730) (1048576,30900)};
\node[font=\tiny, red, anchor=south] at (axis cs:65536,80000) {OOM};
\end{groupplot}
\end{tikzpicture}
\caption{V4-Flash indexer scaling on a single H200, log-log axes. The materialize path runs at $S{=}32{,}768$ (red dot, 129~GB) and OOMs at $S{=}65{,}536$ (red ``X'', plotted at 200~GB and 80~s as visual placeholders for the off-chart OOM). \textsc{StreamIndex}'s chunked path (blue) scales linearly to $S{=}1{,}048{,}576$ at 6.21~GB peak HBM. The 32$\times$ regime extension is the horizontal span between the OOM marker and the rightmost blue point.}
\label{fig:scaling}
\end{figure}

Table~\ref{tab:scaling} is the headline result; Figure~\ref{fig:scaling} plots both axes. The materialize indexer OOMs at $S{=}65{,}536$: the $[B, S, H_I, T]$ FP32 intermediate is 256~GB at this size, exceeding the 140~GB HBM of an H200. The chunked indexer runs to $S{=}1{,}048{,}576$ with 6.21~GB peak. The runnable-$S$ regime extends from 32K (where materialize uses 129/140~GB) to 1M (chunked, 6.21/140~GB) — a 32$\times$ extension on a single H200.

At $S{=}32$K where both fit, the chunked path is also faster (10.3$\times$). This is because the materialize path produces a 64~GB intermediate before the head-sum, then a separate 1~GB intermediate after. The Triton score kernel does the head-sum inline and never allocates the large intermediate, so even when memory is not the binding constraint the chunked path wins on memory-traffic.

\subsection{V4-Pro dimensions}
\begin{table}[t]
\centering
\caption{V4-Pro dimensions: $H_I{=}64$, $d_h{=}128$, $k{=}1024$, $d{=}7168$. Same regime extension pattern as V4-Flash, with peak HBM roughly $2\times$ at the same $S$ due to $k$ doubling.}
\label{tab:v4pro}
\footnotesize
\setlength{\tabcolsep}{3pt}
\begin{tabular}{@{}rrrrrrr@{}}
\toprule
$S$ & $T$ & mat (ms) & mat HBM & chunk (ms) & chunk HBM & speedup \\
\midrule
32{,}768 & 8{,}192 & 317.9 & 129.00 GB & 32 & 0.64 GB & 9.9$\times$ \\
65{,}536 & 16{,}384 & OOM & --- & 127 & 1.02 GB & $\infty$ \\
131{,}072 & 32{,}768 & OOM & --- & 502 & 1.77 GB & $\infty$ \\
262{,}144 & 65{,}536 & OOM & --- & 2{,}004 & 3.27 GB & $\infty$ \\
524{,}288 & 131{,}072 & OOM & --- & 7{,}994 & 6.27 GB & $\infty$ \\
1{,}048{,}576 & 262{,}144 & OOM & --- & 31{,}973 & 12.27 GB & $\infty$ \\
\bottomrule
\end{tabular}
\end{table}

The V4-Pro indexer step (Table~\ref{tab:v4pro}) shows the same OOM threshold at $S{=}65{,}536$ for materialize and the same 1M chunked ceiling. The peak HBM is roughly $2\times$ that of V4-Flash because $k{=}1024$ doubles the running-buffer allocations.

\subsection{Pipeline composition with TileLang attention}

The chunked indexer is one component of a CSA forward; it must compose cleanly with whichever attention kernel runs downstream. We test composition with TileLang's pipelined sparse attention kernel~\cite{tilelang}, the production-quality CUDA reference for V4 attention.

\begin{table}[t]
\centering
\caption{Full V4-Flash pipeline with TileLang's pipelined attention kernel as the common backend, comparing materialize indexer vs chunked indexer. ``HBM'' is the full pipeline peak. Materialize OOMs at $S{=}262{,}144$; chunked runs.}
\label{tab:composition}
\footnotesize
\setlength{\tabcolsep}{3pt}
\begin{tabular}{@{}rrrrrr@{}}
\toprule
$S$ & mat (ms) & mat HBM & chunk (ms) & chunk HBM & verdict \\
\midrule
8{,}192 & 6.5 & 0.58 GB & 7.5 & 0.58 GB & both ok \\
32{,}768 & 42.6 & 2.32 GB & 47.2 & 2.32 GB & both ok \\
65{,}536 & 132.9 & 4.64 GB & 152.3 & 4.64 GB & both ok \\
131{,}072 & 460.4 & \textbf{17.00 GB} & 531.3 & \textbf{9.28 GB} & chunk 1.83$\times$ lower peak \\
\textbf{262{,}144} & \textbf{OOM} & --- & \textbf{1{,}968.8} & \textbf{18.56 GB} & \textbf{chunk only} \\
\bottomrule
\end{tabular}
\end{table}

Table~\ref{tab:composition} confirms the result is not specific to our attention kernel. With TileLang's CUDA attention as the backend, the materialize indexer paired with TileLang attention OOMs at $S{=}262{,}144$ on an H200; the chunked indexer paired with the same TileLang attention runs in 1.97~s. The chunked indexer adds 10--15\% wall-clock overhead at $S$ where the materialize path also runs, in exchange for $S$ scaling past the materialize OOM point (we measured to $S{=}1{,}048{,}576$) and a 1.83$\times$ lower peak HBM at $S{=}131{,}072$.

\subsection{Design-space sweep}

\begin{table}[t]
\centering
\caption{Design-space sweep at $S{=}262{,}144$, V4-Flash dimensions. Recall measured at $S{=}16{,}384$ (where materialize fits as ground truth). Times in milliseconds; HBM in GB peak.}
\label{tab:sweep}
\footnotesize
\setlength{\tabcolsep}{3pt}
\begin{tabular}{@{}lccccc@{}}
\toprule
\multicolumn{6}{l}{\textbf{$c_S$ sweep} ($c_T{=}8192, k{=}512$)} \\
\midrule
$c_S$ & 1024 & 4096 & 16384 & 65536 & 262144 \\
time at S=256K & 4540 & 4165 & 4028 & 4030 & 4030 \\
HBM at S=256K & 1.61 & 1.92 & 3.19 & 8.25 & 28.50 \\
\midrule
\multicolumn{6}{l}{\textbf{$c_T$ sweep} ($c_S{=}2048, k{=}512$)} \\
\midrule
$c_T$ & 1024 & 4096 & 16384 & 65536 & 262144$^{\dagger}$ \\
time at S=256K & 9482 & 4999 & 3900 & 3377 & \textbf{1607} \\
HBM at S=256K & 1.58 & 1.63 & 1.87 & 2.81 & 2.81 \\
\midrule
\multicolumn{6}{l}{\textbf{$k$ sweep} ($c_S{=}2048, c_T{=}8192$)} \\
\midrule
$k$ & 64 & 256 & 512 & 1024 & 2048 \\
time at S=256K & 1754 & 1881 & 1934 & 1996 & 2217 \\
HBM at S=256K & 0.35 & 0.93 & 1.71 & 3.27 & 6.38 \\
\bottomrule
\end{tabular}

\smallskip
\footnotesize
$^{\dagger}$\,$c_T{=}262{,}144$ is clamped to $T{=}65{,}536$, i.e., a single $T$-tile per $S$-tile.
\end{table}

Table~\ref{tab:sweep} reports three sweeps. Across all 15 cells the mean set-overlap recall rounds to 1.0000; min recall is at least 0.9980 in every cell.

The $c_S$ sweep shows a flat time floor (about 4~s) at $c_S \ge 4096$ with peak HBM growing linearly. Below $c_S{=}1024$, Python launch overhead dominates and time rises.

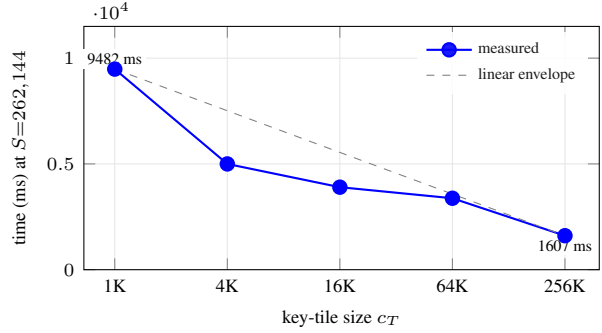
\begin{figure}[t]
\centering
\begin{tikzpicture}
\begin{axis}[
  width=8.4cm, height=4.8cm,
  xmode=log, log basis x=2, xlabel={key-tile size $c_T$},
  xmin=700, xmax=400000,
  xtick={1024,4096,16384,65536,262144},
  xticklabels={1K,4K,16K,64K,256K},
  xticklabel style={font=\scriptsize}, yticklabel style={font=\scriptsize},
  label style={font=\scriptsize}, title style={font=\scriptsize},
  ylabel={time (ms) at $S{=}262{,}144$},
  ymin=0, ymax=11500,
  legend style={font=\tiny, draw=none, fill=white, fill opacity=0.9,
                at={(0.98,0.98)}, anchor=north east, cells={anchor=west}},
  grid=both, grid style={gray!20},
]
\addplot[mark=*, thick, blue, mark size=2.5pt] coordinates {(1024,9482) (4096,4999) (16384,3900) (65536,3377) (262144,1607)};
\addplot[mark=none, dashed, gray] coordinates {(1024,9482) (262144,1607)};
\node[font=\tiny, fill=white, inner sep=1.5pt, anchor=south] at (axis cs:1024,9482) {9482 ms};
\node[font=\tiny, fill=white, inner sep=1.5pt, anchor=north] at (axis cs:262144,1607) {1607 ms};
\legend{measured, linear envelope}
\end{axis}
\end{tikzpicture}
\caption{Key-tile size sweep at $S{=}262{,}144$, V4-Flash dimensions ($c_S{=}2048$, $k{=}512$, $T{=}65{,}536$). Going from $c_T{=}1024$ (64 $T$-tiles) to $c_T{=}T$ (a single $T$-tile per $S$-tile) drops wall-clock by $5.9\times$ at modest memory cost (peak HBM rises from 1.58 to 2.81~GB). Larger key-tile is uniformly better when memory permits.}
\label{fig:sweep}
\end{figure}

The $c_T$ sweep is the most actionable finding (Figure~\ref{fig:sweep}): \textbf{larger $c_T$ is monotonically faster across the measured range on H200}, and the fastest configuration is a single $T$-tile per $S$-tile (i.e., $c_T = T$). Per-tile launch overhead and the merge step both scale with the tile count, while the per-tile compute is well-amortized across HBM bandwidth at any $c_T$ above 4K. The right rule for picking chunk size is: \textit{set $c_T$ as large as fits in the $c_S \cdot c_T \cdot 4$~bytes budget, then pick $c_S$ to control the score-tile size}.

The $k$ sweep is roughly linear in HBM and weakly sublinear in time across $k \in \{64, 256, 512, 1024, 2048\}$. The persistent output buffer (top-$k$ indices and scores) accounts for most of the HBM growth.

\subsection{Ablations}

\begin{table}[t]
\centering
\caption{Ablations. Recall measured at $S{=}16{,}384$, $T{=}4096$ (V4-Flash dims, $k{=}512$); wall-clock measured at $S{=}262{,}144$. All recall ablations use $c_T{=}1024$ to put the chunked driver in the multi-chunk regime over $T$ (4 tiles); A2 uses $c_T{=}256$ to trigger the saturated branch ($c_T < k$). ``A2 ctrl'' is the production chunked path at $c_T{=}256$, isolating the saturated-chunk skip from the small-$c_T$ effect.}
\label{tab:ablation}
\footnotesize
\setlength{\tabcolsep}{4pt}
\begin{tabular}{@{}lcccrr@{}}
\toprule
variant & mean rec. & min rec. & \% = 1 & time (ms) & HBM (GB) \\
\midrule
production (FP32, full merge) & 1.0000 & 0.9980 & 99.99 & 4311.9 & 1.58 \\
A1 no per-chunk merge & \textbf{0.5957} & \textbf{0.1914} & \textbf{25.03} & --- & --- \\
A2 skip saturated ($c_T{=}256$) & \textbf{0.0002} & \textbf{0.0000} & \textbf{0.02} & --- & --- \\
A2 ctrl: production at $c_T{=}256$ & 1.0000 & 0.9980 & 99.99 & --- & --- \\
A3 FP16 score accumulation & 0.9998 & 0.9941 & 91.82 & 3523.7 & 1.07 \\
\bottomrule
\end{tabular}
\end{table}

Table~\ref{tab:ablation} reports three ablations against the production path.

\textbf{A2 (skip saturated chunks)} sets $c_T{=}256 < k{=}512$ and {\it skips} chunks rather than including all entries (the production behaviour). Recall drops to 0.02\%. The production path's branch that includes all entries when $c_T < k$ is essential: those small-$T$ tiles contain top-$k$ entries that materialize would have selected.

\textbf{A2 control} is the same $c_T{=}256$ in the production path, recovering full mean recall (1.0000) and min recall $\ge$ 0.9980. The drop in A2 is attributable to the skip, not to the small chunk size itself.

\textbf{A3 (FP16 score accumulation)} casts the per-tile score and the running buffer to FP16. Mean recall stays at 0.9998 (essentially identical) but the fraction of perfect-recall rows drops from 99.99\% to 91.82\%: roughly 8\% of rows lose a small number of entries to FP16 boundary errors. At the same $c_T{=}1024$ multi-chunk config, the wall-clock at $S{=}262{,}144$ drops to 3.52~s (1.22$\times$ speedup over production's 4.31~s) and peak HBM drops to 1.07~GB (1.48$\times$ lower than production's 1.58~GB). FP16 is a viable precision/throughput tradeoff in inference settings where exact top-$k$ is not required; FP32 is necessary for bit-exact parity with the reference.

\section{Discussion and Limitations}

\paragraph{No real-model end-to-end result (primary limitation).}
We do not load V4-Flash checkpoint weights. The weights would be 270~GB at FP8~\cite{fp8}, exceeding a single H200, and weight offloading would make the run I/O-bound and uninformative for the indexer-step claim. We have not run the chunked indexer through retrieval (needle-in-a-haystack, RULER~\cite{ruler}), QA (LongBench~\cite{longbench}), or code (SWE-bench-Verified~\cite{swebench}) tasks at long context, and we have not measured perplexity drift against materialize. The kernel-level evidence we present is necessary but not sufficient for a "drop-in" production claim; multi-GPU follow-up is required to close that gap. We narrow ``drop-in'' to mean \emph{the chunked indexer step is bit-exact at the layer level} and reserve full-pipeline behavior preservation as future work.

\paragraph{Synthetic input distributions.}
Inputs in our parity tests match per-element variance of the post-projection pipeline but not the trained-checkpoint structure. The bit-exact set-match result is an algorithmic claim (the chunked indexer is the same function as materialize+topk on the same inputs) and does not depend on input distribution; the recall-at-design-sweep finding does.

\paragraph{Tie-order parity is set-level, not order-level.}
Our chunked driver and the materialize reference both call \texttt{torch.topk}, whose tie-breaking is unspecified. The partition-merge invariance is therefore only required to hold at the set level. We verify bit-exact set match against the materialize reference on every parity test (Table~\ref{tab:parity}). For a strict order guarantee, the algorithm needs the lexicographic comparator $\succ$ inlined; we did not implement that because the downstream attention kernel takes the index list as a set.

\paragraph{The chunked driver re-reads inputs.}
Total HBM I/O is the same as materialize: every $q, K_C, w$ entry is read $\lceil S/c_S \rceil$ or $\lceil T/c_T \rceil$ times, the same as the materialize einsum. Only peak is reduced. At $S$ where materialize fits, materialize is faster (at $S{=}8192$, the materialize-path pipeline composition is 6.5~ms vs the chunked-path 7.5~ms). The chunked path is the only viable option at long context, not a uniformly faster path.

\paragraph{Triton attention is not production-quality at $d_h{=}512$.}
Our auxiliary Triton attention forward at V4-Pro dimensions reaches only 8\% of the H200 BF16 peak (versus TileLang's 40\% peak with the same shape). The Triton compiler in version 3.7 does not expose the FlashAttention-3 (FA3)-style~\cite{flashattn3} warp-specialization patterns needed to close that gap, which would require a CUDA reimplementation of the attention step. The contribution of this paper is on the indexer step; the attention is orthogonal and we deliberately compose with TileLang for the production-quality path.

\paragraph{Future work.} (i) End-to-end V4-Flash inference with weight offloading or multi-GPU tensor parallelism (TP), on retrieval (needle-in-a-haystack), QA (LongBench), and code (SWE-bench-Verified) benchmarks; (ii) a fused indexer-attention kernel sharing $K_C$ loads across both stages; (iii) multi-hardware validation on H100, A100, and MI300X to confirm the regime extension is hardware-agnostic.

\section{Conclusion}

The lightning indexer in V4-class CSA gates the pipeline at long context: with $H_I{=}64$ heads and ratio $m{=}4$, the materialize-then-topk reference path OOMs at $S{=}65{,}536$ with V4-Flash dimensions on a single H200. \textsc{StreamIndex} ships a chunked partition-merge top-$k$ driver over a fused Triton score kernel that scales the same indexer step to $S{=}1{,}048{,}576$ with 6.21~GB peak HBM, a 32$\times$ regime extension. The chunked driver runs unmodified with TileLang's pipelined attention kernel and adds 10--15\% wall-clock at sub-OOM $S$. Bit-exact parity with the materialize path is verified at small $S$. Code: \url{https://github.com/RightNow-AI/StreamIndex}.

\end{document}